\documentclass[bookreview]{template/clv2}

\usepackage{url}
\usepackage[nolist]{acronym}
\usepackage{framed}

\runningtitle{The Influence of Context on Dialog Act Recognition}

\runningauthor{Ribeiro et al.}

\begin{document}

\title{The Influence of Context on Dialog Act Recognition}

\author{Eug\'{e}nio Ribeiro\thanks{L$^2$F {--} Spoken Language Systems Lab {--} INESC-ID Lisboa, R. Alves Redol, 9, 1000-029 Lisboa, Portugal Instituto~Superior~T\'{e}cnico,~Universidade de Lisboa, Av. Rovisco Pais, 1, 1049-001 Lisboa, Portugal E-mail:~eugenio.ribeiro@l2f.inesc-id.pt}}
\affil{INESC-ID Lisboa \\ Instituto Superior T\'{e}cnico, Universidade de Lisboa}

\author{Ricardo Ribeiro\thanks{L$^2$F {--} Spoken Language Systems Laboratory {--} INESC-ID Lisboa, Portugal Instituto~Universit\'{a}rio~de~Lisboa~(ISCTE-IUL), Av. das For\c{c}as Armadas, 1649-026 Lisboa, Portugal E-mail:~ricardo.ribeiro@inesc-id.pt}}
\affil{INESC-ID Lisboa \\ Instituto Universit\'{a}rio de Lisboa (ISCTE-IUL)}

\author{David Martins de Matos\thanks{L$^2$F {--} Spoken Language Systems Lab {--} INESC-ID Lisboa, R. Alves Redol, 9, 1000-029 Lisboa, Portugal Instituto~Superior~T\'{e}cnico,~Universidade de Lisboa, Av. Rovisco Pais, 1, 1049-001 Lisboa, Portugal E-mail:~david.matos@inesc-id.pt}}
\affil{INESC-ID Lisboa \\ Instituto Superior T\'{e}cnico, Universidade de Lisboa}

\maketitle

%
%

\begin{acronym}
    \acro{SVM}{Support Vector Machine}
    \acro{HMM}{Hidden Markov Model}
    \acro{NLU}{Natural Language Understanding}
    \acro{BN}{Bayesian Network}
    \acro{DBN}{Dynamic Bayesian Network}
    \acro{MLP}{Multi-Layer Perceptron}
    \acro{CU-Boulder}{University of Colorado Boulder}
    \acro{CMU}{Carnegie Mellon University}
    \acro{CRF}{Conditional Random Field}
    \acro{IVR}{Interactive Voice Response}
    \acro{SMO}{Sequential Minimal Optimization}
    \acro{FCT}{Funda\c{c}\~{a}o para a Ci\^{e}ncia e a Tecnologia}
    \acro{ASR}{Automatic Speech Recognition}
    \acro{WER}{Word Error Rate}
\end{acronym}

%
%

%
%

\begin{abstract}
This article presents an analysis of the influence of context information on dialog act recognition. We performed experiments on the widely explored Switchboard corpus, as well as on data annotated according to the recent ISO 24617-2 standard. The latter was obtained from the Tilburg DialogBank and through the mapping of the annotations of a subset of the Let's Go corpus. We used a classification approach based on SVMs, which had proved successful in previous work and allowed us to limit the amount of context information provided. This way, we were able to observe the influence patterns as the amount of context information increased. Our base features consisted of n-grams, punctuation, and wh-words. Context information was obtained from one to five preceding segments and provided either as n-grams or dialog act classifications, with the latter typically leading to better results and more stable influence patterns. In addition to the conclusions about the importance and influence of context information, our experiments on the Switchboard corpus also led to results that advanced the state-of-the-art on the dialog act recognition task on that corpus. Furthermore, the results obtained on data annotated according to the ISO 24617-2 standard define a baseline for future work and contribute for the standardization of experiments in the area.  
\end{abstract}

%
%

%
%

\section{Introduction}
\label{sec:introduction}

As \namecite{Searle1969} stated, dialog, speech, or illocutionary acts are the minimal units of linguistic communication, as they reveal the intention behind the uttered words. Thus, automatic dialog act recognition is an important task in \ac{NLU}, as by identifying the intention of the conversational partner the interpretation process is simplified. This is particularly important for the development of more robust and natural dialog systems, since many communication problems that occur with existing systems are due to misinterpretations of ambiguous utterances, which could be disambiguated if the intention was correctly identified.   

During a conversation, the communicative intention of a speaker and, thus, the uttered words, depends on the current state of the dialog~\cite{Stone2002}. For instance, if the conversation is just starting, the speaker will probably utter a greeting due to politeness conventions. Also, if the conversational partner asked a question, the speaker will probably intend to answer it, although, in some cases, he or she may choose to just simply ignore it for some reason. Nonetheless, either way, this means that dialog acts are influenced by the state and context of the dialog. However, the speaker's communicative intention is not usually related to the fact that the conversational partner asked a question ten turns before. This suggests that the range of influence of previous utterances is limited, or, at least, that the influence of a given utterance decays as the conversation evolves.

In this article, we study the influence of context information, extracted from previous segments in multiple ways, on dialog act recognition. With our experiments, we assess the importance of such information for the task, its range of influence, and the best ways to represent it.

Some dialog act recognition approaches (Section~\ref{ssec:recognition}) try to predict the best dialog act sequence for a whole dialog and, thus, rely not only on past and present information, but also on future information to classify each segment. Although such approaches have applications, they are not useful for live interactions, since the system does not have access to future information at the time it needs to assess the intention of its conversational partner. Thus, our studies do not explore future information and rely only on information extracted from the current and past segments.

The remaining sections of this article are organized as follows: Section~\ref{sec:related} provides related work on dialog act recognition by presenting an overview of existing annotated data, as well as describing multiple approaches and state-of-the-art results for different corpora. Section~\ref{sec:corpora} describes the datasets used in our experiments. Section~\ref{sec:setup} defines our experimental setup by describing the used features, classification approach, and evaluation methodology. Section~\ref{sec:results} presents a comparative evaluation of the obtained results, both among the different approaches and datasets, and with the state-of-the-art. Finally, the achieved results are discussed in Section~\ref{sec:discussion} and directions for future work are presented in Section~\ref{sec:future}.

%
%

%
%

\section{Related Work}
\label{sec:related}

Dialog act recognition is a classification task that attributes a dialog act label to each dialog segment. In this sense, multiple classification approaches have been applied to this task. To our knowledge, all of them were supervised approaches. This means that large amounts of annotated data are required to obtain solid models. Thus, corpora selection plays an important role in the task. In terms of evaluation, previous studies relied solely on accuracy as performance measure. Below, we analyze different annotated corpora and classification approaches that were previously applied on the task.

\subsection{Data Annotation}
\label{ssec:annotation}

Multiple corpora have been annotated in terms of dialog acts. Table~\ref{tab:corpora} presents some of those corpora and their characteristics. We can see that multiple domains, languages, and kinds of interaction are covered, which enables portability experiments and domain and interaction-independent conclusions. However, on the other hand, the used tag sets are not standardized among corpora. While some of the corpora, such as DCIEM Map Task~\cite{Bard1995} and AMI Meeting~\cite{Carletta2005}, were annotated using reduced tag sets under 20 tags, others, such as Dihana~\cite{Benedi2006} and NESPOLE~\cite{Costantini2002}, were annotated using tag sets with hundreds of tags. Furthermore, while some {--} DCIEM Map Task, Switchboard~\cite{Jurafsky1997}, SCHISMA~\cite{Keizer2002b}, ICSI-MRDA~\cite{Shriberg2004}, and AMI Meeting {--} were annotated using domain-independent tag sets that can be used for annotating any corpora, others {--} VERBMOBIL~\cite{Kay1994}, NESPOLE, Dihana, and LEGO~\cite{Schmitt2012} {--} were annotated using domain-dependent tag sets, which are limited to corpora in that domain. This means that the tag sets were developed with different objectives and have different hierarchies and levels of abstraction, which makes cross-corpora and generalization experiments hard to perform. Also, while some of the corpora, such as Switchboard, ICSI-MRDA, and AMI Meeting, have several tens of thousands of annotated segments, others, like SCHISMA, have less than a thousand.

\begin{table}[htbp]
    \caption{Characteristics of some corpora annotated in terms of dialog acts. All corpora have at least one human speaker in each dialog, thus, the Interaction column refers to the nature of the other speaker. The last column, DD, states whether the tag set is domain-dependent or not.}
    \label{tab:corpora}
    \begin{tabular}{l r r r r r r}
        \hline
        Corpus         & Interaction & Domain    & Language & Segments & \#Tags & DD  \tabularnewline
        \hline
        VERBMOBIL      & Human       & Schedules & Multiple &    58961 &     72 & YES \tabularnewline
        DCIEM Map Task & Human       & Routes    & English  &     4787 &     12 & NO  \tabularnewline
        Switchboard    & Human       & Open      & English  &   223606 &     44 & NO  \tabularnewline
        SCHISMA        & Wizard      & Theatre   & Dutch    &      440 &     64 & NO  \tabularnewline
        NESPOLE        & Human       & Tourism   & Multiple &    12565 &   1168 & YES \tabularnewline
        ICSI-MRDA      & Human       & Meetings  & English  &   105000 &     55 & NO  \tabularnewline
        AMI Meeting    & Human       & Meetings  & English  &   102198 &     15 & NO  \tabularnewline
        Dihana         & Wizard      & Trains    & Spanish  &    23008 &    248 & YES \tabularnewline
        LEGO           & Machine     & Buses     & English  &    14186 &     50 & YES \tabularnewline
    \end{tabular}
\end{table}

In an attempt to standardize dialog act annotation and, thus, set the ground for more comparable research in the area, \namecite{Bunt2012} defined the ISO 24617-2 standard. The first thing that should be noted in the standard is that annotations should be performed on functional segments rather than on turns or utterances~\cite{Carroll1978}. This should happen because a single turn or utterance may have multiple functions, revealing different intentions. However, automatic functional segmentation is a complex task on its own. Thus, according to the standard, dialog act annotation does not consist of a single label, but rather of a complex structure containing information about the participants, relations with other functional segments, the semantic dimension of the dialog act, its communicative function, and optional qualifiers concerning certainty, conditionality, partiality, and sentiment. In terms of semantic dimensions, the standard defines nine {--} \em Task\em, \em Auto-Feedback\em, \em Allo-Feedback\em, \em Turn Management\em, \em Time Management\em, \em Discourse Structuring\em, \em Own Communication Management\em, \em Partner Communication Management\em, and \em Social Obligations Management\em. Communicative functions are equivalent to the dialog act labels present in the multiple tag sets used to annotate the corpora presented in Table~\ref{tab:corpora}. They were divided into general-purpose functions, which can occur in any semantic dimension, and dimension-specific functions, which, as the name indicates, are specific to a certain dimension. The set of general-purpose functions is hierarchically distributed according to Table~\ref{tab:generaldistr}. Dimension-specific functions are all at the same level and are distributed across dimensions according to Table~\ref{tab:dimensiondistr}. This means that the \em Task \em dimension contains general-purpose functions only.

\begin{table}[htbp]
    \caption{Distribution of general-purpose functions according to the ISO 24617-2 standard.}
    \label{tab:generaldistr}
    \begin{tabular}{l r}
        \hline
        Function              & Count \tabularnewline
        \hline
        Information-seeking   &     4 \tabularnewline
        Information-providing &     6 \tabularnewline
        Commissive            &     4 \tabularnewline
        Directive             &     5 \tabularnewline
    \end{tabular}
\end{table}

\begin{table}[htbp]
    \caption{Distribution of dimension-specific functions according to the ISO 24617-2 standard.}
    \label{tab:dimensiondistr}
    \begin{tabular}{l r}
        \hline
        Dimension                        & Count \tabularnewline
        \hline
        Auto-feedback                    &     2 \tabularnewline
        Allo-feedback                    &     3 \tabularnewline
        Turn Management                  &     6 \tabularnewline
        Time Management                  &     2 \tabularnewline
        Discourse Structuring            &     2 \tabularnewline
        Own Communication Management     &     2 \tabularnewline
        Partner Communication Management &     3 \tabularnewline
        Social Obligations Management    &    10 \tabularnewline
    \end{tabular}
\end{table}

\subsection{Dialog Act Recognition}
\label{ssec:recognition}

Existing approaches for dialog act recognition can be split into two categories. The ones that try to predict the best dialog act sequence for a given set of segments and the ones that predict the dialog act of each segment individually. The approaches in the first category take advantage of algorithms such as \acp{CRF}~\cite{Lafferty2001}, \acp{HMM}~\cite{Baum1966}, and other \acp{BN}~\cite{Friedman1997}. On the other hand, approaches in the second category take advantage of algorithms such as Neural Networks~\cite{McCulloch1988}, Decision Trees~\cite{Breiman1984}, and \acp{SVM}~\cite{Cortes1995}. We could organize related work according to these two categories. However, since experiments are spread among multiple corpora, it would be difficult to compare the different approaches. Thus, we opted to organize related work by corpora, meaning that experiments performed on, at least, similar data are presented together, allowing easier comparison.

Switchboard is probably the most explored corpus for the dialog act recognition task. However, multiple variations of the original 44-label tag set have been used, differing mainly on how abandoned, unrecognized, and interrupted segments are dealt with. Thus, the number of tags varies between 41 and 44. The first experiments on this corpus were performed by \namecite{Stolcke2000}, using word n-grams as features for an \ac{HMM} and a 42-label variant of the tag set. Using manual transcriptions, the best result, 71.0\% accuracy, was obtained using trigrams. This value decreased to 64.8\% when using automatic transcriptions. The same authors also used Decision Trees based on prosodic information, but only achieved 49.7\% accuracy. Later, \namecite{Rotaru2002} used a Memory-Based Learning approach~\cite{Aha1991} to obtain 72.32\% accuracy, also using the 42-label variant of the tag set. He used the k-NN algorithm~\cite{Cover1967} with the distance between neighbors being measured as the number of common bigrams between utterances, according to a hash function. \namecite{Sridhar2009} used a maximum entropy model combining lexical, syntactic, and prosodic features, as well as context information extracted from the three previous segments. In terms of lexical and syntactic features, the authors used word n-grams, POS tags, and Supertags, the enriched descriptions of lexical items proposed by \namecite{Bangalore1999}. As for acoustic-prosodic features, they used pitch, energy, and accent and boundary tone labels. Context information was provided by extracting the same features from the surrounding segments, as well as in the form of the dialog act labels for those segments. The experiments were performed using the 42-label variant of the tag set, as well as a compressed version with 7 labels. Using the 42-label variant, the authors achieved 70.4\% accuracy without context information, and 76.0\% when it was included from preceding segments. These values decreased to 55.1\% and 59.7\% when automatically recognized segments were used instead of manual transcriptions. Using the compressed version, the results increased to 82.5\% and 83.1\%, and 69.9\% and 73.9\%, respectively. The authors also performed experiments using information extracted from the next three segments, achieving 71.3\% and 56.1\% on the 42-label variant, and 82.8\% and 70.7\% on the compressed version. \namecite{Webb2010} were able to achieve 80.72\% accuracy by applying a classification approach based on cue phrases, that is, phrases that are highly indicative of a particular dialog act. However, they used a 41-label variant of the tag set, merging different kinds of statement into a class covering 49\% of the corpus. Finally, \namecite{Gamback2011} used \acp{SVM}, together with an active learning approach to select the most informative subset of the training data, to obtain 76.50\%, 76.34\%, and 77.85\% accuracy on the 42, 43, and 44-label variants of the tag set, respectively. The used features included multiple n-grams, punctuation, and wh-words, as well as some context information in the form of n-grams from the previous segments.

On the DCIEM Map Task Corpus, experiments were performed by \namecite{Wright1998} using three different approaches with similar results. The combination of \acp{HMM} and an intonation model achieved 64\% accuracy, while Decision Trees trained with the CART algorithm~\cite{Breiman1984} achieved 63\% accuracy. Additionally, a \ac{MLP}~\cite{Rosenblatt1962} with one hidden layer, with suprasegmental and prosodic features based on duration as inputs achieved 62\% accuracy. \namecite{Sridhar2009} also performed experiments on this corpus, using the same approach described for the Switchboard corpus. However, in this case, only manual transcriptions were used. They achieved 66.6\% without context information, 72.5\% using information from the previous segments, and 67.4\% using information from the following segments. 

The NESPOLE corpus was explored by \namecite{Levin2003}. The presence or absence of grammar characteristics {--} 212 for English and 259 for German {--} was used as a set of binary features for four different classification approaches. Memory-Based Learning, through the application of the IB1 algorithm~\cite{Aha1991}, achieved 69.82\% accuracy for English and 67.57\% for German. Decision Trees trained with the C4.5 algorithm~\cite{Quinlan1993} achieved 70.41\% for English and 67.90\% for German. A \ac{MLP} achieved 71.52\% for English and 67.61\% for German. Finally, a Naive Bayes~\cite{Friedman1997} classifier achieved 51.39\% for English and 46.00\% for German. By appending word bigram information to the Memory-Based Learning approach, accuracy increased to 81.25\% for English and 78.93\% for German. 

Several other corpora were explored for dialog act recognition using a single approach. For instance, \namecite{Samuel1998} were able to achieve 71.22\% accuracy on the VERBMOBIL corpus, annotated with the 18-label domain independent subset of the original tag set. For that, they used Transformation-Based Learning~\cite{Brill1995} with a Monte Carlo strategy~\cite{Metropolis1949}. On the SCHISMA corpus, \namecite{Keizer2002} used \acp{BN} with sentence type, subject type, and punctuation as features to achieve 44\% accuracy. Using a switching \ac{DBN}~\cite{Ghahramani1998} with a trigram language model and prosodic features, \namecite{Dielmann2007} obtained 60\% accuracy on the AMI Meeting corpus. Following the same direct classification approach based on cue phrases applied to the Switchboard corpus, \namecite{Webb2010} achieved 58.14\% accuracy on the ICSI-MRDA corpus. Finally, \namecite{Gamback2011} also used the \ac{SVM}-based approach applied to the Switchboard corpus on the Dihana corpus. They performed experiments using the 248-label tag set, as well as a domain-independent subset with 72 tags, obtaining 90.97\% and 94.08\% accuracy, respectively.

Overall, we can see that experiments on dialog act recognition have been widely spread both in terms of approaches and corpora. This makes it difficult to compare results, even for experiments on the same corpora, since different tag sets and evaluation procedures have been used. Still, on the most explored corpus, Switchboard, the \ac{SVM} approach used by \namecite{Gamback2011} seems to surpass the other approaches.

In terms of features, lexical features, especially n-grams, are the most used. However, acoustic-prosodic features have also been used, generally in experiments that did not involve textual information. Other features, such as sentence and subject type, are hard to obtain automatically and are themselves indicative of the dialog act. Thus, their identification can be seen as an intermediate step towards dialog act recognition.

Finally, since we want to assess the influence of context information on dialog act recognition in the context of a dialog system, it is important to notice that approaches that predict the best dialog act sequence for a whole dialog or even ones that take everything that happened since the beginning of the dialog into account when classifying a given segment are not indicated. This is true for two reasons. First, some of those approaches rely on future information, that is, they use information not available to a dialog system at the time of classification, to classify a given segment. The second reason is that when such approaches, it is hard to limit the amount of provided context information, making it difficult to control the analysis we want to perform. Thus, a non-sequential classification approach, to which context information can be provided in the form of different features which provide that sequential information, is more indicated.     

%
%

%
%

\section{Corpora}
\label{sec:corpora}

ISO 24617-2 is the current and only existent standard for dialog act annotation. However, since it is a recent standard, the amount of data annotated according to it is small, which leaves room for questions regarding the solidity of the results achieved by experiments performed on it. Thus, we also performed experiments on the large and widely explored Switchboard corpus.  

%
%
\subsection{Switchboard}
\label{ssec:switchboard}

Switchboard~\cite{Godfrey1992} is a corpus consisting of about 2400 telephone conversations among 543 American English speakers (302 male and 241 female). Each pair of speakers was automatically attributed a topic for discussion, from 70 different ones. Furthermore, speaker pairing and topic attribution were constrained so that no two speakers would be paired with each other more than once and no one spoke more than once on a given topic. Speech from the two subjects was recorded into separate channels, using an 8 kHz sampling rate.

A subset of 1155 manually transcribed conversations (annotated with disfluency, abandonment, and interruption information), containing 223606 segments, was annotated using the SWBD-DAMSL tag set~\cite{Jurafsky1997}. Dialog act annotation was performed by eight Linguistics graduate students at \ac{CU-Boulder} during a three-month period. The SWBD-DAMSL tag set was structured so that the annotators were able to label the conversations from transcriptions alone. It contains over 200 unique tag combinations. However, in order to obtain higher inter-annotator agreement and higher example frequencies per class, a less fine-grained set of 44 tags was devised. The class distribution (Table~\ref{tab:sbdistr}) is highly unbalanced, with the three most frequent classes {---} \em Statement-opinion \em (36\%), \em Acknowledgement \em (19\%), and \em Statement-non-opinion \em (13\%) {---} covering 68\% of the corpus. The set can be reduced to 43 or 42 categories~\cite{Stolcke2000,Rotaru2002,Gamback2011}, if the \em Abandoned \em and \em Uninterpretable \em categories are merged, and depending on how the \em Segment \em category (used when the current segment is the continuation of the previous one by the same speaker) is treated. By analyzing the data, we came to the conclusion that merging segments labeled as \em Segment \em with the previous segment by the same speaker is the best approach, because some of the attributed labels only made sense when the segments were merged. Also, it makes sense to merge the \em Abandoned \em and \em Uninterpretable \em categories, because both represent disruptions in the dialog flow, which interfere with the typical dialog act sequence. However, in our experiments, we used the three variants of the tag set to allow direct comparison with the related work. There is also a 41-category variant of the tag set~\cite{Webb2010}, which merges the \em Statement-opinion \em and \em Statement-non-opinion \em categories, making the most frequent class cover 49\% of the corpus.

\begin{table}[htbp]
    \caption{Label distribution in the Switchboard Dialog Act Corpus (Replicated from~\protect\cite{Jurafsky1997}).}
    \label{tab:sbdistr}
    \begin{tabular}{l r r | l r r}
        \hline
        Label                   & Count &  \%  & Label                 & Count &  \%  \tabularnewline
        \hline
        Statement-non-opinion   & 72824 &  36  & Collab Completion     &   699 &  .4  \tabularnewline
        Acknowledgement         & 37096 &  19  & Repeat-Phrase         &   660 &  .3  \tabularnewline
        Statement-opinion       & 25197 &  13  & Open-Question         &   632 &  .3  \tabularnewline
        Agreement               & 10820 &   5  & Rhetorical-Question   &   557 &  .3  \tabularnewline
        Abandoned               & 10569 &   5  & Hold                  &   540 &  .2  \tabularnewline
        Appreciation            &  4663 &   2  & Reject                &   338 &  .2  \tabularnewline
        Yes-No-Question         &  4624 &   2  & Neg Non-no Answer     &   292 &  .1  \tabularnewline
        Non-verbal              &  3548 &   2  & Non-understanding     &   288 &  .1  \tabularnewline
        Yes Answer              &  2934 &   1  & Other Answer          &   279 &  .1  \tabularnewline
        Conventional Closing    &  2486 &   1  & Conventional Opening  &   220 &  .1  \tabularnewline
        Uninterpretable         &  2158 &   1  & Or-Clause             &   207 &  .1  \tabularnewline
        Wh-Question             &  1911 &   1  & Dispreferred Answers  &   205 &  .1  \tabularnewline
        No Answer               &  1340 &   1  & 3rd-party-talk        &   115 &  .1  \tabularnewline
        Response Acknowledge    &  1277 &   1  & Offers / Options      &   109 &  .1  \tabularnewline
        Hedge                   &  1182 &   1  & Self-talk             &   102 &  .1  \tabularnewline
        Decl-Yes-No-Question    &  1174 &   1  & Downplayer            &   100 &  .1  \tabularnewline
        Other                   &  1074 &   1  & Maybe                 &    98 & \textless.1  \tabularnewline
        Backchannel-Question    &  1019 &   1  & Tag-Question          &    93 & \textless.1  \tabularnewline
        Quotation               &   934 &  .5  & Decl-Wh-Question      &    80 & \textless.1  \tabularnewline
        Summarize               &   919 &  .5  & Apology               &    76 & \textless.1  \tabularnewline
        Aff Non-yes Answer      &   836 &  .4  & Thanking              &    67 & \textless.1  \tabularnewline
        Action Directive        &   719 &  .4  &                       &       &              \tabularnewline
    \end{tabular}
\end{table}

This subset is called the Switchboard Dialog Act Corpus but is referred to simply as Switchboard in this article. Figure~\ref{fig:sbdialog} shows an excerpt of one of the transcriptions, where each line corresponds to an annotated segment. \namecite{Stolcke2000} describe a data partition of this subset into a training set of 1115 conversations, a test set of 19 conversations, and a future use set of 21 conversations. However, the concrete partition is not disclosed and, thus, in the remaining related bibliography there is no reference to this partition and cross-validation is used for evaluation.

\begin{figure}[htbp]
    \begin{framed}
    \begin{dialogue}
        Speaker A: Okay. /

        Speaker A: \{D So, \}

        Speaker B: [ [ I guess, +

        Speaker A: What kind of experience [ do you, + do you ] have, then with child care?

        Speaker B: I think, ] + \{F uh, \} I wonder ] if that worked. /

        Speaker A: Does it say something? /

        Speaker B: I think it usually does. /

        Speaker B: You might try, \{F uh, \} /

        Speaker B: I don't know, /

        Speaker B: hold it down a little longer, /

        Speaker B: \{C and \} see if it, \{F uh, \} -/

        Speaker A: Okay \textless beep\textgreater. /
    \end{dialogue}
    \end{framed}
    \caption{An excerpt of a Switchboard corpus transcription. Brackets are used to annotate different phenomena. Square brackets signal repetitions and corrections. Curly brackets signal disfluencies.}
    \label{fig:sbdialog}
\end{figure}

We selected this corpus for our experiments because it contains a large amount of annotated data, which can lead to solid results. Furthermore, it has been widely explored, which allows result comparison with previous work. Finally, its tag set is domain-independent, which reduces the probability of drawing conclusions that depend on the domain of the corpus.

%
%
\subsection{ISO 24617-2 Data}
\label{ssec:iso}

As stated in Section~\ref{sec:related}, the ISO 24617-2 standard defines guidelines for dialog act annotation, including communicative functions in multiple dimensions, dependencies between dialog acts, and modifiers concerning, for instance, conditionality and partiality. Not all of these aspects are relevant for our studies. In fact, although it could be interesting to analyze the influence of context information for all dimensions, only the task dimension has enough diversity to be analyzed using the same procedure as on the Switchboard corpus. Thus, in the studies presented in this article, we only explore the influence of context information in the recognition of communicative functions in the task dimension.

We decided to perform experiments on data annotated according to this standard in an attempt to contribute to the uniformization of research on dialog acts. However, since the standard is relatively new, not much data has been annotated according to its tag set. In this sense, we were only able to obtain the data provided by the Tilburg DialogBank~\cite{Bunt2016}~\footnote{\url{https://dialogbank.uvt.nl/}}. Thus, to obtain more data, we decided to look into other annotated datasets whose annotations could be mapped into the ones of the standard. There are attempts at converting other annotation formats into the standard. For instance, the SWBD-DAMSL used to annotate the Switchboard corpus~\cite{Fang2012}. However, these approaches involve manual steps which are highly time consuming. Thus, we looked into a different corpus, LEGO~\cite{Schmitt2012}, an annotated subset of the Let's Go corpus~\cite{Raux2006}, which has been used in many dialog related tasks and whose domain-dependent dialog act annotations could be mapped into the communicative functions defined by the standard almost directly. Although this is not a complete annotation according to the standard, it provided a large amount of data for our studies in comparison to what we were able to obtain from the DialogBank. More detailed information about the datasets is provided below.

\subsubsection{Tilburg DialogBank}

The Tilburg University DialogBank~\cite{Bunt2016} provides multiple dialogs annotated according to the ISO 24617-2 standard. The dialogs are extracted from different corpora in multiple languages. At the time the studies presented in this article were performed, 11 English dialogs and 7 Dutch dialogs were available in the DialogBank, distributed as shown in Table~\ref{tab:tilburg}. It is important to notice that the amount of available data is small, especially in Dutch. In terms of labels, information providing functions are dominant overall, with the \em inform \em tag being present in around 13\% of the segments.

\begin{table}[htbp]
    \caption{Information about the dialogs obtained from the Tilburg DialogBank.}
    \label{tab:tilburg}
    \begin{tabular}{l l r r p{.3\textwidth}}
        \hline
        Corpus & Language & \#Dialogs & \#Segments & Dominant Tag \tabularnewline
        \hline
        Switchboard    & English & 2  & 554  & inform (36\%)      \tabularnewline
        TRAINS         & English & 3  & 236  & inform (19\%)      \tabularnewline
        HCRC Map Task  & English & 6  & 2095 & instruct (14\%)    \tabularnewline
        DIAMOND        & Dutch   & 3  & 88   & inform (14\%)      \tabularnewline
        Dutch Map Task & Dutch   & 1  & 93   & inform (19\%)      \tabularnewline
        OVIS           & Dutch   & 3  & 91   & answer (12\%)      \tabularnewline
        \hline
        All            & English & 11 & 2885 & inform (13\%)      \tabularnewline
        All            & Dutch   & 7  & 272  & inform (14\%)      \tabularnewline
    \end{tabular}
\end{table}

\subsubsection{LEGO}
 
The LEGO corpus~\cite{Schmitt2012} is an annotated subset of the The Let's Go corpus~\cite{Raux2006}. It features data from the \ac{CMU}'s Let's Go Bus Information System, which provides information about bus schedules in the city of Pittsburg, through spoken telephonic interaction with a dialog system.

A set of 347 calls recorded during 2006 was later annotated by the Dialogue Systems Group at Ulm University, Germany. This annotated corpus is the one usually referred to as LEGO. It features 14186 utterances {--} 9083 system utterances and 5103 user utterances. The number of utterances is different because some system turns contain multiple utterances and some of its prompts are not answered by the user. There is no standard division of this corpus into training and testing sets.

Since system utterances are generated through slot filling of fixed templates, they have no errors and contain casing and punctuation information. In contrast, the transcriptions of user utterances were obtained using an \ac{ASR} system and, thus, contain no casing, nor punctuation information. Furthermore, the recognition was not always correct, due to the \ac{WER} of the \ac{ASR} system, for which a concrete value is not revealed. Figure~\ref{fig:legodialog} presents an excerpt of a transcription. 

\begin{figure}[htbp]
    \begin{framed}
    \begin{dialogue}
        System: Welcome to the CMU Let's Go bus information system.
        
        System: To get help at any time, just say Help or press zero.
        
        System: What can I do for you?
        
        User: PENN AT BUTLER TIME IS IT IS THE CENTRE AT HIGHLAND FROM RUNS THE 61A EIGHT THIRTY 61B CAN'T OAKLAND
        
        System: Leaving from OAKLAND . Is this correct?
        
        User: OAKLAND
        
        System: Leaving from OAKLAND . Is this correct?
        
        User: YES
        
        System: Right.
        
        System: Where do you want to go?
        
        User: SQUIRREL HILL OF FIFTH AVENUE
        
        System: Going to FIFTH AVENUE . Is this correct?
        
        User: YES
    \end{dialogue}
    \end{framed}
    \caption{An excerpt of a LEGO corpus transcription.}
    \label{fig:legodialog}
\end{figure}

In terms of dialog acts, the LEGO corpus was originally annotated using two distinct and domain-dependent tag sets for system an user turns. The set for system turns contained 28 tags, such as \em Ask Destination\em, \em Ask Confirm Bus\em, and \em Ask Time\em, while the set for user turns contained 22 tags, such as \em Place Information\em, \em Confirm Destination\em, and \em Reject Bus\em. When using such tags, context information is clearly very important for dialog act recognition, since a given dialog act drastically reduces the number of non-disruptive possibilities, that is, that do not break the dialog flow, for the next one. Exploring dialog act recognition under these conditions is not relevant for our study. However, most of these domain-dependent tags can be directly mapped into ISO 24617-2 communicative functions. Thus, in order to obtain more data annotated according to the standard, we performed that mapping as described in~\cite{Ribeiro2016}. We did not take some of the dimensions into account, as the transcriptions did not contain enough information to allow annotations relative to those dimensions. However, since our study focuses on the \em Task \em dimension, information about those dimensions is not relevant. This way, we obtained over 4 times the number of annotated segments we were able to obtain from the Tilburg DialogBank. The label distribution across the corpus is presented in Table~\ref{tab:legodistr}, both for the whole corpus and considering system and user turns separately. In this sense, the nature of the corpus is highly noticeable in the difference between system and user turns, with the system using mainly questions and instructions and the user answering those questions.

\begin{table}[htbp]
    \caption{Label distribution in the LEGO corpus.}
    \label{tab:legodistr}
    \begin{tabular}{l r r r r r r}
        \hline
                      & \multicolumn{2}{c}{All} & \multicolumn{2}{c}{System} & \multicolumn{2}{c}{User} \tabularnewline
        Label         & Count & \% & Count & \% & Count & \% \tabularnewline
        \hline
        Check Question &  2257 & 16 &  2256 & 25 &     1 & \textless.1 \tabularnewline
        Set Question   &  2197 & 16 &  1987 & 22 &   210 &  4 \tabularnewline
        Instruct       &  1918 & 14 &  1812 & 20 &   106 &  2 \tabularnewline
        Answer         &  1462 & 10 &     0 &  0 &  1462 & 29 \tabularnewline
        Inform         &  1256 &  9 &   656 &  7 &   600 & 12 \tabularnewline
        Confirm        &  1162 &  8 &     0 &  0 &  1162 & 23 \tabularnewline
        Disconfirm     &  1105 &  8 &     0 &  0 &  1105 & 22 \tabularnewline
        Promise        &   277 &  2 &   277 &  3 &     0 &  0 \tabularnewline
        Request        &   155 &  1 &    70 &  1 &    85 &  2 \tabularnewline
        Suggest        &    40 & .3 &    40 & .4 &     0 &  0 \tabularnewline
    \end{tabular}
\end{table}

%
%

%
%

\section{Experimental Setup}
\label{sec:setup}

We approached dialog act recognition as a supervised classification task, following the typical steps for this kind of task. This section describes our options in terms of feature selection, classification approaches, and evaluation methodologies.

%
%

\subsection{Features}
\label{ssec:features}

Dialog acts are related to language and, consequently, to the words present in each utterance, as well as to the intonation of those utterances. This means that both textual and audio features are important to recognize dialog acts, as was shown in some of the studies presented in Section~\ref{sec:related}~\cite{Wright1998,Stolcke2000,Dielmann2007,Sridhar2009}. However, for the experiments presented in this document, we relied just on lexical features extracted from conversation transcripts. We opted for this approach since lexical features have been widely used and proved efficient in the related work. This means that the effort of obtaining alignments between the audio and transcriptions of all segments is unnecessary for studying context influence patterns. Nonetheless, we believe that audio features would be able to improve the overall classification results and, thus, they should be explored as future work.

The transcripts were subjected to a normalization step, consisting of converting all words to lowercase, appending tokens signaling start and end of each segment, and separating punctuation from the words.

\subsubsection{Base Features}

We used the frequencies of word n-grams as the main features extracted from the current segment, as n-grams are able to capture keywords and word sequence information. In order to select which n-grams to use, we performed experiments using a specific n-gram length, between 1 and 5, as well as using a cumulative \em n\em, also between 1 and 5. Since \acp{SVM} were used to obtain the best results on the Switchboard corpus~\cite{Gamback2011}, we used an \ac{SVM} classifier with only n-gram frequencies as features for these experiments. Table~\ref{tab:ngrams} presents the results obtained on the 42-label Switchboard. We can see that using both unigrams and bigrams led to the best results. However, along each row, the result differences are not statistically significant. Nonetheless, for \em n \em larger than 2, there is statistical significance in the difference between the results obtained using a specific \em n \em and a cumulative \em n \em. This means that the information provided by unigrams and bigrams is relevant. Thus, the remaining experiments presented in this article used unigrams and bigrams as features.

\begin{table}[htbp]
    \caption{Accuracy (\%) results obtained on the 42-label Switchboard corpus using an SVM classifier with n-gram frequencies as features. The first row presents results obtained using a specific n-gram length, while the second presents results obtained using a cumulative \em n\em , that is, using all n-grams with \em n \em between 1 and the pivot of the column.}
    \label{tab:ngrams}
    \begin{tabular}{l c c c c c}
        \hline
        & \multicolumn{5}{c}{n} \tabularnewline
        & 1 & 2 & 3 & 4 & 5     \tabularnewline
        \hline
        Specific n   & 72.69 & \textbf{73.03} & 72.07 & 69.64 & 65.60 \tabularnewline
        Cumulative n & 72.69 & \textbf{73.69} & 73.36 & 73.26 & 73.20 \tabularnewline
    \end{tabular}
\end{table}

In addition to n-grams, we also used the existence of wh-words and punctuation as features. The first provides important cues for question detection and the last may disambiguate different intentions behind the same words. For instance, exclamation marks may turn a statement into a command, while the placement of commas may change the whole meaning and intention of a sentence.

\subsubsection{Context Features}

Since the focus of this article is the influence of context information on dialog act recognition, we used two different approaches to capture such information. The first one uses the n-grams extracted from the preceding segments as features of the segment being classified, while the second uses the dialog act classification of the preceding segments instead. While the first approach focuses on the sequence of words and sentences, the second focuses on the sequence of intentions. Furthermore, the first approach can be separated into two different approaches. One that uses the n-grams from the preceding segments directly and another that tags those n-grams with an index corresponding to the distance, in number of segments, between the segment they where extracted from and the current segment. The first considers all n-grams equally and, thus, focuses only on word sequences, while the second distinguishes the n-grams according to their origin and, thus, also considers sentence sequences and relative distances.  

In order to assess the range of influence, we performed experiments using context information extracted from the \em n \em preceding segments, with \em n \em between 1 and 5.

%
%

%
%

\subsection{Classification}
\label{ssec:classification}

As stated in Section~\ref{sec:related}, multiple dialog act recognition approaches, such as the one applied by \namecite{Stolcke2000} on the Switchboard corpus, try to predict the best dialog act sequence for a given conversation. However, our work focuses on dialog act recognition during a conversation between a dialog system and its conversational partner. In this scenario such approaches are not useful as, although it may have expectations, the system has no way to be certain of how the conversation will evolve. Thus, it must only rely on previous and current information. Furthermore, since we want to assess the influence of context information on dialog act recognition, we must be able to limit the amount of information provided. Thus, instead of a sequential approach, such as \acp{HMM} or \acp{CRF}, we opted for an approach based on \acp{SVM}, which have already been used on the state-of-the-art approach for this task on the Switchboard corpus~\cite{Gamback2011}.

Due to the large number of features and the size of the Switchboard corpus, we opted for using the linear kernel, with 0.1 as the value of the cost parameter. For the experiments on the Switchboard corpus, we used LIBLINEAR~\cite{Fan2008} to train the classifiers, since it is well-suited to deal with large amounts of data. For the experiments on corpora annotated according to the ISO 24617-2 standard, we took advantage of the \ac{SMO} algorithm~\cite{Platt1998} implementation provided by the Weka Toolkit~\cite{Hall2009}.

%
%

%
%

\subsection{Evaluation}
\label{ssec:evaluation}

We use accuracy, that is, the ratio between the number of correct predictions and the total number of predictions, as the performance measure, since it has been consistently chosen as the measure to evaluate performance in dialog act recognition.

Since there are no fully disclosed training and testing partitions of the corpora, a strict comparative study is not possible. Thus, we opted for using 10-fold cross-validation as the evaluation procedure. However, in order to perform comparisons with some of the related work, other numbers of folds were also used. Nonetheless, unless otherwise stated, the presented results were obtained using 10-fold cross-validation.

In order to assess the statistical significance of the differences between the multiple results obtained on the same data, we defined a significance level of 5\% and performed the Wilcoxon Signed-Rank Test~\cite{Wilcoxon1945}. Thus, in this article, when we say that some difference is significant/insignificant, it means that the \em p\em-value of the test was below/above 5\%.  

%
%

%
%

%
%

\section{Results}
\label{sec:results}

This section presents the results we obtained on the Switchboard corpus and data annotated according to the ISO 24617-2 standard, using different approaches to provide context information. As in Section~\ref{sec:related}, we present the results grouped by corpus to facilitate the comparison between the multiple approaches. However, some cross-corpora remarks are also performed along this section and discussed in Section~\ref{sec:discussion}. 

\subsection{Switchboard}

As stated in Section~\ref{sec:corpora}, we performed experiments using three variants of the SWBD-DAMSL tag set, with 42 to 44 tags. Tables~\ref{tab:sb42results},~\ref{tab:sb43results}, and~\ref{tab:sb44results} show the results obtained using the 42, 43, and 44-label tag sets, respectively. The first thing that should be noticed is that the baseline, that is, the accuracy result obtained without context information, is above 70\% for every tag set variant {--} 73.69\%, 70.59\%, and 70.57\%, respectively.

\begin{table}[htbp]
    \caption{Accuracy (\%) results obtained on the 42-label Switchboard corpus using context information extracted from the \em n \em preceding segments in different forms. The first two rows refer to context information provided in the form of n-grams while the last refers to context information provided in the form of dialog act classifications.}
    \label{tab:sb42results}
    \begin{tabular}{l c c c c c c}
        \hline
        & \multicolumn{6}{c}{\# Previous Segments} \tabularnewline
        & 0 & 1 & 2 & 3 & 4 & 5 \tabularnewline
        \hline
        Untagged N-Grams            & \textbf{73.69} & 57.72 & 49.99 & 45.88 & 42.83 & 40.54 \tabularnewline
        Index-Tagged N-Grams        & 73.69 & \textbf{74.92} & 74.18 & 73.65 & 73.28 & 73.27 \tabularnewline
        Dialog Act Labels           & 73.69 & 78.20 & 78.88 & \textbf{79.06} & 79.03 & 79.03 \tabularnewline
    \end{tabular}
\end{table}

\begin{table}[htbp]
    \caption{Accuracy (\%) results obtained on the 43-label Switchboard corpus using context information extracted from the \em n \em preceding segments in different forms. The first two rows refer to context information provided in the form of n-grams while the last refers to context information provided in the form of dialog act classifications.}
    \label{tab:sb43results}
    \begin{tabular}{l c c c c c c}
        \hline
        & \multicolumn{6}{c}{\# Previous Segments} \tabularnewline
        & 0 & 1 & 2 & 3 & 4 & 5 \tabularnewline
        \hline
        Untagged N-Grams            & \textbf{70.59} & 57.45 & 51.10 & 44.66 & 41.18 & 38.52 \tabularnewline
        Index-Tagged N-Grams        & 70.59 & 72.98 & \textbf{75.16} & 74.78 & 74.49 & 74.44 \tabularnewline
        Dialog Act Labels           & 70.59 & 75.55 & 76.21 & \textbf{76.38} & \textbf{76.38} & 76.36 \tabularnewline
    \end{tabular}
\end{table}

\begin{table}[htbp]
    \caption{Accuracy (\%) results obtained on the 44-label Switchboard corpus using context information extracted from the \em n \em preceding segments in different forms. The first two rows refer to context information provided in the form of n-grams while the last refers to context information provided in the form of dialog act classifications.}
    \label{tab:sb44results}
    \begin{tabular}{l c c c c c c}
        \hline
        & \multicolumn{6}{c}{\# Previous Segments} \tabularnewline
        & 0 & 1 & 2 & 3 & 4 & 5 \tabularnewline
        \hline
        Untagged N-Grams            & \textbf{70.57} & 57.52 & 51.09 & 44.48 & 40.96 & 38.24 \tabularnewline
        Index-Tagged N-Grams        & 70.57 & 72.97 & \textbf{75.12} & 74.76 & 74.49 & 74.41 \tabularnewline
        Dialog Act Labels           & 70.57 & 75.56 & 76.29 & \textbf{76.42} & 76.40 & 76.36 \tabularnewline
    \end{tabular}
\end{table}

By looking at the rows corresponding to the context information provided in the form of untagged n-grams, we can see that it is detrimental, considerably decreasing accuracy for every tag set. Furthermore, the accuracy result significantly decreases as the number of preceding segments increases. This phenomenon can be explained by the fact that Switchboard dialogs do not have a fixed domain and, thus, except for social obligations, the occurrence of similar segment sequences is relatively rare throughout the corpus. Furthermore, since the dialogs have long segments, the addition of n-grams from preceding segments without distinguishing them from the ones of the current segment ends up giving more weight to the previous segments than to the current one in terms of n-gram frequencies. This makes consecutive segments have similar frequencies in spite of having different classifications. Furthermore, similarity increases as \em n \em increases, since only the most distant segment is replaced by the new segment. This ends up reducing entropy among the different segments and, consequently, impairing the classification process and explaining the results.

On the other hand, context information provided in the form of index-tagged n-grams was able to significantly improve the baseline between 1.23 and 4.57 percentage points. This happened because index tags provide additional sequence information and effectively distinguish n-grams extracted from different segments, making each segment more distinct from the others and, thus, increasing entropy in an important way for a classification task with a large number of classes. However, the influence pattern seems to be different for the multiple variants of the tag set. We can see that information extracted from the first preceding segment is able to significantly improve accuracy for every variant, between 1.23 and 2.40 percentage points. However, for the 42-label variant, information extracted from additional segments significantly reduces accuracy, even to values below the baseline after the second preceding segment. On the other hand, information extracted from the second preceding segment is still able to significantly improve accuracy by an additional 2.18 and 2.15 percentage points for the 43 and 44-label variants, respectively. Beyond that, accuracy starts decreasing, but never gets below the baseline, nor even below the accuracy obtained using context information extracted from a single preceding segment.

The last row of the tables concerns context information provided in the form of dialog act classifications, that is, the labels attributed to the preceding segments. We can see that by appending the classification of a single preceding segment, accuracy significantly increased for every tag set variant, between 4.51 and 4.99. Approximately an additional percentage point can be added to this value by appending information from additional segments, until the results start to stabilize. However, the improvements provided by preceding segments beyond the second are not statistically significant for any tag set variant. Furthermore, for the 42-label variant, even the improvement provided by the second preceding segment is not statistically significant. This reinforces the importance of the first preceding segment and suggests that the influence of context information highly decreases with the distance between segments.

\begin{figure}[htbp]
    \includegraphics[width=\textwidth]{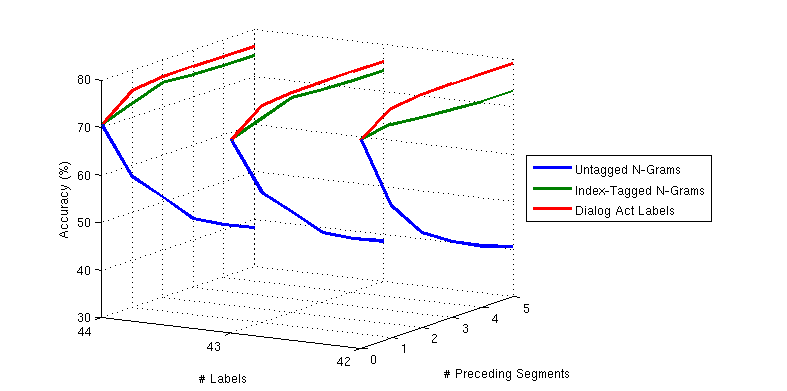}
    \caption{Accuracy (\%) results obtained on the Switchboard corpus using context information extracted from the \em n \em preceding segments in different forms.}
    \label{fig:swbd}
\end{figure}

Figure~\ref{fig:swbd} shows that the approach that used dialog act labels as context information surpassed the ones that used n-grams for every tag set variant. However, the labels used were the manual annotations of the corpus and, thus, the obtained results are an upper bound for the approach. In order to assess the performance of this approach without relying on gold standard annotations, we performed experiments using automatic classifications. To obtain the automatic classifications, we split the corpus in half and trained classifiers without context information on different subsets of the corpus and used them to predict the labels for the second half. We used three different subsets {--} the second half, the whole corpus, and the first half {--} to assess the impact of the dependence between the training and evaluation sets, from complete dependence when using the second half to train to complete independence when using the first half. The accuracy of the labels is presented in Table~\ref{tab:nocontextclass}. As expected, the accuracy when using an independent set to train is much lower than when using a dependent set.

\begin{table}[htbp]
    \caption{Accuracy (\%) results on the second half of the Switchboard corpus obtained by classifiers without context information trained using different subsets of the corpus.}
    \label{tab:nocontextclass}
    \begin{tabular}{l c c c}
        \hline
        & \multicolumn{3}{c}{\# Labels} \tabularnewline
        & 42 & 43 & 44 \tabularnewline
        \hline
        Second Half  & 86.88 & 85.45 & 85.20 \tabularnewline
        Whole Corpus & 85.30 & 83.70 & 83.76 \tabularnewline
        First Half   & 71.53 & 69.59 & 69.73 \tabularnewline
    \end{tabular}
\end{table}

\begin{figure}[htbp]
    \includegraphics[width=\textwidth]{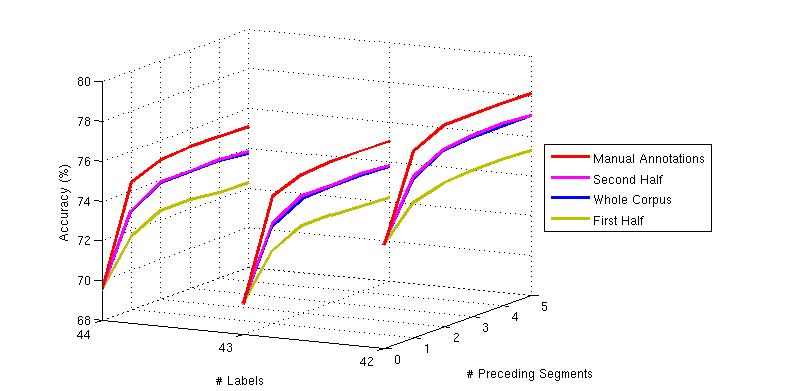}
    \caption{Accuracy (\%) results obtained on the second half of the Switchboard corpus using context information extracted from the \em n \em preceding segments in the form of dialog act labels.}
    \label{fig:swbdauto}
\end{figure}

In order to assess the performance when using the automatically obtained labels as context information, we trained classifiers on the half of the corpus for which they were predicted. We also trained classifiers on that data using the manual annotations to assess the decrease in accuracy. The 10-fold cross-validation results obtained by these classifiers are presented in Tables~\ref{tab:sb42labels},~\ref{tab:sb43labels}, and~\ref{tab:sb44labels}, respectively. It is interesting to notice that the decrease in accuracy of the classifiers without context information in relation to the ones trained on the whole corpus was of just 0.49 percentage points for the 42-label tag set variant and 1.08 and 1.04 for the 43 and 44-label variants, respectively. In terms of the classifiers using context information, the first thing that should be noticed, and that can also be seen in Figure~\ref{fig:swbdauto}, is that, as expected, the accuracy of the classifiers decreased as the accuracy of the labels used to provide context information decreased. Furthermore, it is important to notice that the influence patterns observed when using manual annotations remained the same when using automatically obtained labels. In this sense, the first preceding segment was always the most informative, with the following providing smaller and smaller amounts of additional information, which typically led to accuracy increments without statistical significance. Talking about statistical significance, there is no significance between the results obtained when using the labels predicted by the classifier trained on the whole corpus and the ones predicted by the classifier trained on the second half. Furthermore, for the 43 and 44-label tag set variants, there is also no statistical significance between the results obtained using those labels and the ones obtained using manual annotations. However, the decrease in accuracy when using the labels predicted by the classifier trained on the first half of the corpus is always significant. In this sense, the decrease in relation to when using manual annotations was of 2.81 percentage points for the 42-label variant, 2.68 for the 43-label variant, and 2.65 for the 44-label variant.

\begin{table}[htbp]
    \caption{Accuracy (\%) results obtained on the second half of the 42-label Switchboard corpus using context information extracted from the \em n \em preceding segments in the form of dialog act labels.}
    \label{tab:sb42labels}
    \begin{tabular}{l c c c c c c}
        \hline
        & \multicolumn{6}{c}{\# Previous Segments} \tabularnewline
        & 0 & 1 & 2 & 3 & 4 & 5 \tabularnewline
        \hline
        Manual Annotations  & 73.20 & 77.37 & 78.08 & 78.13 & \textbf{78.15} & 78.12 \tabularnewline
        Second Half         & 73.20 & 76.10 & 76.88 & 77.05 & \textbf{77.14} & 77.05 \tabularnewline
        Whole Corpus        & 73.20 & 75.99 & 76.83 & 76.94 & \textbf{77.00} & 76.99 \tabularnewline
        First Half          & 73.20 & 74.80 & 75.21 & 75.31 & \textbf{75.34} & 75.26 \tabularnewline
    \end{tabular}
\end{table}
 
\begin{table}[htbp]
    \caption{Accuracy (\%) results obtained on the second half of the 43-label Switchboard corpus using context information extracted from the \em n \em preceding segments in the form of dialog act labels.}
    \label{tab:sb43labels}
    \begin{tabular}{l c c c c c c}
        \hline
        & \multicolumn{6}{c}{\# Previous Segments} \tabularnewline
        & 0 & 1 & 2 & 3 & 4 & 5 \tabularnewline
        \hline
        Manual Annotations  & 69.51 & 74.39 & 74.96 & \textbf{75.08} & 75.07 & 75.06 \tabularnewline
        Second Half         & 69.51 & 73.01 & 73.89 & 73.88 & \textbf{73.96} & 73.83 \tabularnewline
        Whole Corpus        & 69.51 & 72.87 & 73.72 & 73.81 & \textbf{73.84} & 73.76 \tabularnewline
        First Half          & 69.51 & 71.65 & 72.40 & \textbf{72.40} & 72.29 & 72.23 \tabularnewline
    \end{tabular}
\end{table}

\begin{table}[htbp]
    \caption{Accuracy (\%) results obtained on the second half of the 44-label Switchboard corpus using context information extracted from the \em n \em preceding segments in the form of dialog act labels.}
    \label{tab:sb44labels}
    \begin{tabular}{l c c c c c c}
        \hline
        & \multicolumn{6}{c}{\# Previous Segments} \tabularnewline
        & 0 & 1 & 2 & 3 & 4 & 5 \tabularnewline
        \hline
        Manual Annotations  & 69.53 & 74.42 & 75.00 & 75.11 & \textbf{75.12} & 75.09 \tabularnewline
        Second Half         & 69.53 & 72.96 & 73.89 & 73.90 & \textbf{73.97} & 73.83 \tabularnewline
        Whole Corpus        & 69.53 & 72.92 & 73.84 & 73.87 & \textbf{73.91} & 73.76 \tabularnewline
        First Half          & 69.53 & 71.69 & 72.46 & \textbf{72.47} & 72.30 & 72.28 \tabularnewline
    \end{tabular}
\end{table}

\begin{figure}[htbp]
    \includegraphics[width=\textwidth]{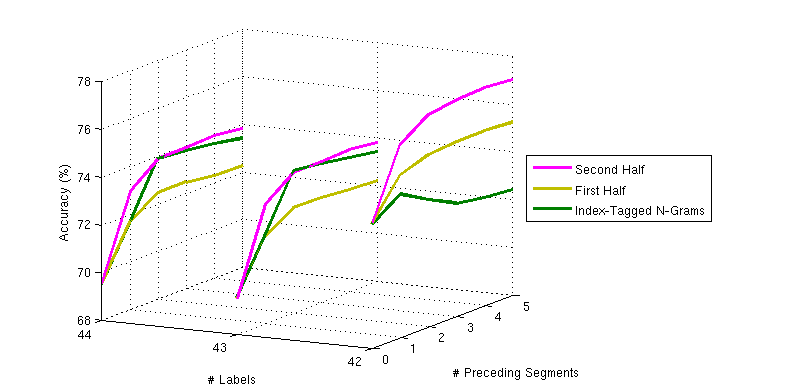}
    \caption{Accuracy (\%) results obtained on the second half of the Switchboard corpus using context information extracted from the \em n \em preceding segments in the form of automatic dialog act labels and index-tagged n-grams.}
    \label{fig:swbdtng}
\end{figure}

In order to compare the results obtained using context information in the form of automatic dialog act labels with the ones obtained using n-grams, we also trained classifiers on the second half of the corpus using index-tagged n-grams. Table~\ref{tab:sb2halftng} presents the obtained results. We can see that the results follow the same patterns as on the whole corpus, with only one preceding segment providing relevant information for the 42-label tag set variant, while for the other variants the second preceding segment is still able to improve accuracy. In Figure~\ref{fig:swbdtng} we can see that for the 42-label tag set variant the results obtained using index-tagged n-grams were always below the ones obtained using automatic dialog act labels, even when they were obtained using a classifier trained on the first half of the corpus. On the other hand, for the other variants, the results are around 1.50 percentage points above the ones obtained using automatic labels predicted by a classifier trained on the first half of the corpus. Furthermore, although the results are still below the ones obtained using automatic labels predicted by a classifier trained on the second half of the corpus, that difference is not statistically significant.

\begin{table}[htbp]
    \caption{Accuracy (\%) results obtained on the second half of the Switchboard corpus using context information extracted from the \em n \em preceding segments in the form of index-tagged n-grams.}
    \label{tab:sb2halftng}
    \begin{tabular}{l c c c c c c}
        \hline
        & \multicolumn{6}{c}{\# Previous Segments} \tabularnewline
        & 0 & 1 & 2 & 3 & 4 & 5 \tabularnewline
        \hline
        42 Labels & 73.20 & \textbf{73.99} & 73.33 & 72.73 & 72.54 & 72.42 \tabularnewline
        43 Labels & 69.51 & 71.66 & \textbf{73.95} & 73.82 & 73.63 & 73.43 \tabularnewline
        44 Labels & 69.53 & 71.64 & \textbf{73.92} & 73.78 & 73.64 & 73.42 \tabularnewline
    \end{tabular}
\end{table}
 
Overall, on the Switchboard corpus, we can see that the results obtained using the 43 and 44-label tag set variants are very similar. However, they differ from the ones obtained using the 42-label variant, even in terms of the observable influence patterns as the amount of context information increased. This means that the way in which the \em Segment \em category is treated is relevant for the task. In this sense, as stated in Section~\ref{ssec:switchboard}, we believe that using the 42-label tag set variant, merging segments labeled as \em Segment \em with the previous segment by the same speaker is the best approach. Nonetheless, there are still conclusions which can be drawn independently of the tag set variant. Context information provided in the form of dialog act labels is potentially more informative than in the form of n-grams, especially untagged ones, which have a negative impact on accuracy. We used the term potentially as it depends on the accuracy of those labels. However, for the 42-label tag set variant, that was true for all the experiments using automatically generated labels. Furthermore, the first preceding segment was clearly the most informative both when using index-tagged n-grams and dialog act labels as context features. Beyond that, results obtained using index-tagged n-grams were irregular for the different tag set variants. While for the 42-label tag set variant accuracy started decreasing after the first preceding segment, for the remaining variants the second previous segment was still informative. On the other hand, the results obtained using dialog act labels, both manual and automatic, followed a pattern that suggests a high decrease of influence in relation to the distance to the segment being classified, with smaller and smaller accuracy increments as the distance increased.

In order to compare our results with previous results on the Switchboard corpus it is important to notice that our normalization step did not take the characteristics of the transcriptions into account. However, the transcriptions of the Switchboard corpus include disfluency, abandonment, and interruption annotations, which were processed in the same manner as the remaining words when n-grams were extracted. By altering the normalization step to take these annotations into account, that is, not splitting them and considering them a single token, we were able to improve the best results to 79.60, 78.00, and 77.90, which is particularly significant for the 43 and 44-label tag set variants. These results surpassed the ones obtained by \namecite{Gamback2011} for every tag set variant. On the 42-label variant, the accuracy improvement exceeded 3 percentage points. However, the results obtained using context information in the form of index-tagged n-grams were lower than the reported in their article using similar information. This suggests that their active learning approach is, in fact, able to improve results. Since the concrete corpus partition used by \namecite{Stolcke2000} is not disclosed in their paper, we performed 50-fold cross-validation to obtain results using the same number of training and testing examples as described and, thus, try to obtain more comparable results. Using this setup, we were able to obtain 79.60\% accuracy, which represents an accuracy improvement exceeding 8 percentage points. In order to compare our results with the ones obtained by \namecite{Webb2010}, we also performed an experiment using the 41-label variant of the tag set, by merging the statement categories. Under these conditions, we obtained 86.50\% accuracy, which represents an improvement of almost 6 percentage points. Taking this comparison into account, we believe that our results are the current state-of-the-art for dialog act recognition on the Switchboard corpus.

\subsection{ISO 24617-2 Data}

The experiments performed on data annotated according to the ISO 24617-2 standard using context information in the form of n-grams, both untagged and index-tagged, were identical to the ones performed on the Switchboard corpus. However, the experiments using context information in the form of dialog act labels differed in two aspects. First, we only used manual annotations. Second, we performed experiments using information about all dimensions, as well as using information related to the \em Task \em dimension only. We did not include experiments using automatic annotations for different reasons according to the source of the data. On the LEGO corpus, the system is always aware of the dialog act it produced. Thus, it would be unrealistic to automatically predict those as well. On the other hand, by predicting only the user dialog acts, the experiment would be inherently different from the ones performed on the remaining corpora. On data obtained from the Tilburg DialogBank the main reason is related to its small amount, especially for Dutch, because experiments that split the data even further would drastically reduce accuracy. Furthermore, we would have to automatically produce labels for the remaining dimensions as well. This is problematic since the distribution and nature of the remaining dimensions is completely different from the \em Task \em dimension. Thus, different classifiers or even rule-based approaches would be more indicated to predict dialog acts in those dimensions. Still, we used the manual annotations to perform experiments using context information from all dimensions in an attempt to assess dependencies between the \em Task \em dimension and the others.   

\begin{figure}[htbp]
    \includegraphics[width=\textwidth]{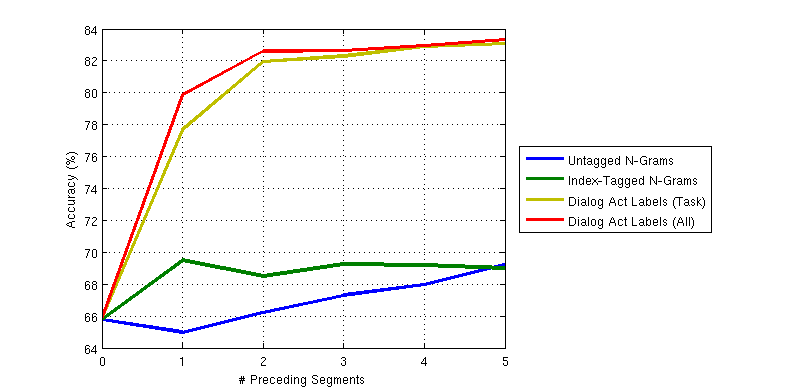}
    \caption{Accuracy (\%) results obtained on the Tilburg DialogBank English dialogs using context information extracted from the \em n \em preceding segments in different forms.}
    \label{fig:banken}
\end{figure}

Concerning data obtained from the Tilburg DialogBank, we performed experiments on both English and Dutch dialogs in order to assess possible language-independent results. Table~\ref{tab:bankenresults} presents the results obtained on English dialogs. We can see that, in this case, context information in the form of untagged n-grams led to much more irregular results than in the case of the Switchboard corpus. Accuracy insignificantly decreases 0.80 percentage points below the baseline when using a single preceding segment but beyond that it starts increasing, from 0.45 percentage points beyond the baseline when using 2 preceding segments up to a significant 3.43 when using 5 preceding segments. However, as shown in Figure~\ref{fig:banken}, this is still the approach with worst performance. For the remaining approaches, the first preceding segment is still the one that leads to the largest accuracy boost. In this sense, the pattern produced by index-tagged n-grams is a mix between the ones produced on the Switchboard corpus. The first preceding segment significantly improves accuracy by 3.71 percentage points and, beyond that, accuracy starts decreasing but never below the baseline. However, these differences are all statistically insignificant. Nonetheless, the 0.28 percentage point difference between the best results obtained using untagged n-grams and tagged n-grams is statistically significant. Finally, the results obtained using dialog act annotations reveal the same influence patterns as on the Switchboard corpus, with a noticeable decrease of influence in relation to the distance to the segment being classified. Furthermore, this approach was also the one that performed better, with a 17.54 percentage point improvement over the baseline, versus the 3.71 percentage points of index-tagged n-grams. Information provided by dimensions other than \em Task \em was able to improve accuracy, but only by 0.25 percentage points. However, if we consider a single preceding segment, the improvement is of 2.18 percentage points, which is more pronounced. Nonetheless, in both cases, the result difference is statistically insignificant.

\begin{table}[htbp]
    \caption{Accuracy (\%) results obtained on the Tilburg DialogBank English dialogs using context information extracted from the \em n \em preceding segments in different forms. The first two rows refer to context information provided in the form of n-grams while the remaining two refer to context information provided in the form of dialog act classifications relative to the \em Task \em dimension only or all the dimensions.}
    \label{tab:bankenresults}
    \begin{tabular}{l c c c c c c}
        \hline
        & \multicolumn{6}{c}{\# Previous Segments} \tabularnewline
        & 0 & 1 & 2 & 3 & 4 & 5 \tabularnewline
        \hline
        Untagged N-Grams            & 65.79 & 64.99 & 66.24 & 67.31 & 67.97 & \textbf{69.22} \tabularnewline
        Index-Tagged N-Grams        & 65.79 & \textbf{69.50} & 68.49 & 69.25 & 69.18 & 69.01 \tabularnewline
        Dialog Act Labels (Task)    & 65.79 & 77.68 & 81.94 & 82.29 & 82.88 & \textbf{83.08} \tabularnewline
        Dialog Act Labels (All)     & 65.79 & 79.86 & 82.60 & 82.63 & 82.95 & \textbf{83.33} \tabularnewline
    \end{tabular}
\end{table}

\begin{figure}[htbp]
    \includegraphics[width=\textwidth]{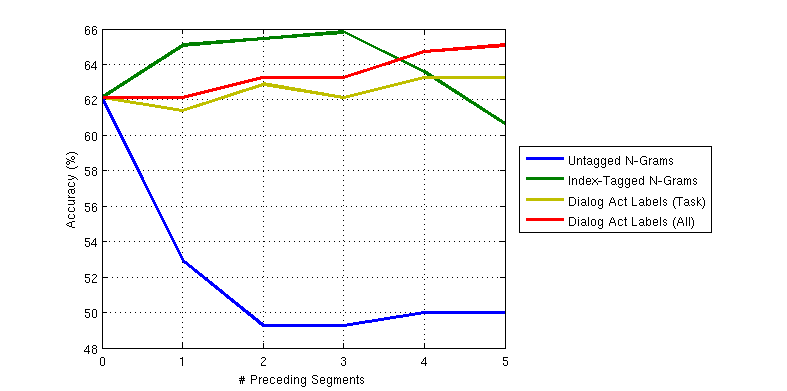}
    \caption{Accuracy (\%) results obtained on the Tilburg DialogBank Dutch dialogs using context information extracted from the \em n \em preceding segments in different forms.}
    \label{fig:bankne}
\end{figure}

The results obtained on Dutch dialogs are presented in Table~\ref{tab:bankneresults}. It shows that, as expected, accuracy results are lower than the ones obtained on English dialogs, since the amount of data is smaller. For the same reason, the influence patterns seem more irregular than in the previous cases, as can be seen in Figure~\ref{fig:bankne}. However, the importance of context information is still noticeable. Contrarily to what happened with the English dialogs, using context information in the form of untagged n-grams followed a detrimental pattern as on the Switchboard corpus. On the other hand, index-tagged n-grams improved accuracy up to the third preceding segment, obtaining the best result on this dataset with 3.68 percentage points over the baseline. However, the first preceding segment was still the most informative, leading to an accuracy improvement of 2.94 percentage points. Beyond the third preceding segment, accuracy started significantly decreasing, even to values below the baseline. Context information provided in the form of dialog act labels was less effective on this dataset, with a maximum improvement of 2.94 percentage points over the baseline. Furthermore, the results did not follow the same pattern as on the previously described experiments. In fact, there is no statistical significance in any of the differences for this approach. This suggests that providing context information in the form of dialog act labels requires larger amounts of training data to be effective in comparison to information provided in the form of index-tagged n-grams. As for information provided by other dimensions, as on the English dialogs, it was also able to slightly improve accuracy, in this case by a maximum of 1.83 percentage points. Given the irregular patterns obtained for Dutch, it is hard to draw language-independent conclusions. However, the importance of context information is noticeable in both cases, as well as the importance of representing context information in a form distinguishable from the information related to the segment being classified, that is, in a way that increases entropy. This is shown in both languages by the improvements provided by index-tagged n-grams and dialog act labels and the detrimental impact of untagged n-grams. Still, we believe that experiments using larger amounts of data in languages other than English could lead to interesting conclusions regarding the language-independence of the influence of context on dialog act recognition.

\begin{table}[htbp]
    \caption{Accuracy (\%) results obtained on the Tilburg DialogBank Dutch dialogs using context information extracted from the \em n \em preceding segments in different forms. The first two rows refer to context information provided in the form of n-grams while the remaining two refer to context information provided in the form of dialog act classifications relative to the \em Task \em dimension only or all the dimensions.}
    \label{tab:bankneresults}
    \begin{tabular}{l c c c c c c}
        \hline
        & \multicolumn{6}{c}{\# Previous Segments} \tabularnewline
        & 0 & 1 & 2 & 3 & 4 & 5 \tabularnewline
        \hline
        Untagged N-Grams            & \textbf{62.13} & 52.94 & 49.26 & 49.26 & 50.00 & 50.00 \tabularnewline
        Index-Tagged N-Grams        & 62.13 & 65.07 & 65.44 & \textbf{65.81} & 63.60 & 60.66 \tabularnewline
        Dialog Act Labels (Task)    & 62.13 & 61.40 & 62.87 & 62.13 & \textbf{63.24} & \textbf{63.24} \tabularnewline
        Dialog Act Labels (All)     & 62.13 & 62.13 & 63.24 & 63.24 & 64.71 & \textbf{65.07} \tabularnewline
    \end{tabular}
\end{table}

The dialogs of the LEGO corpus have a different nature from all the previous ones, since they consist on human-machine interactions. Furthermore, system segments are generated using templates and slot filling. Thus, variations in system segments annotated with the same dialog act label are in small number. This highly impacts accuracy, as can be seen in Tables~\ref{tab:legoresults} and~\ref{tab:legouserresults}, which present the results on the whole corpus and on the user segments only, respectively. The baseline accuracy difference is 14.19 percentage points. Furthermore, in a real situation, the system is aware of all the dialog acts it produced. Thus, it is more interesting to analyze user segments only, that is, the results in Table~\ref{tab:legouserresults}.

\begin{table}[htbp]
    \caption{Accuracy (\%) results obtained on the LEGO corpus using context information extracted from the \em n \em previous segments in different forms. The first two rows refer to context information provided in the form of n-grams while the remaining two refer to context information provided in the form of dialog act classifications relative to the \em Task \em dimension only or all the dimensions.}
    \label{tab:legoresults}
    \begin{tabular}{l c c c c c c}
        \hline
        & \multicolumn{6}{c}{\# Previous Segments} \tabularnewline
        & 0 & 1 & 2 & 3 & 4 & 5 \tabularnewline
        \hline
        Untagged N-Grams            & 91.44 & \textbf{95.36} & 92.11 & 88.66 & 82.36 & 83.58 \tabularnewline
        Index-Tagged N-Grams        & 91.44 & \textbf{95.92} & 95.86 & 95.64 & 95.40 & 95.26 \tabularnewline
        Manual Annotations (Task)   & 91.44 & 94.81 & 95.50 & 95.72 & \textbf{95.83} & 95.78 \tabularnewline
        Manual Annotations (All)    & 91.44 & 95.65 & 95.85 & 95.85 & \textbf{95.94} & 95.85 \tabularnewline
    \end{tabular}
\end{table}

\begin{figure}[htbp]
    \includegraphics[width=\textwidth]{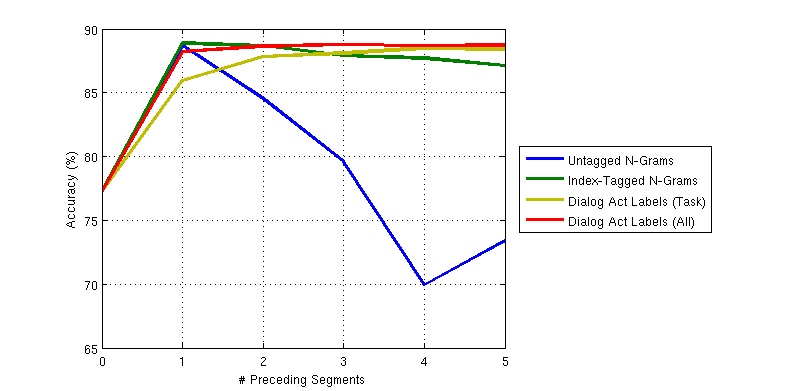}
    \caption{Accuracy (\%) results obtained on the user segments of the LEGO corpus using context information extracted from the \em n \em preceding segments in different forms.}
    \label{fig:legoresults}
\end{figure}

The first thing to notice, and which can also be seen in Figure~\ref{fig:legoresults}, is that the results are much more similar between approaches than on the remaining corpora, with the differences in top results being below 0.50 percentage points and statistically insignificant. This can be explained by the characteristics of the system segments, which improve the accuracy of the approaches that provide context information in the form of n-grams. This becomes even clearer when looking at the patterns produced by appending additional preceding segments. The best result, with an improvement above 10 percentage points over the baseline, is obtained by appending a single segment, which, since we are looking at results for user segments, is a fixed system segment. In these cases, the n-grams extracted from that segment appear multiple times in the corpus, with only a few possible following dialog act labels. Thus, they provide a very important cue for the classifier and accuracy is highly improved. Previous segments beyond the first start to reduce accuracy, as user segments are now taken into account. However, we can still notice that the decrease is much more pronounced for untagged n-grams than for index-tagged n-grams, as the first approach obtains results below the baseline after appending the fourth preceding segment, while the latter still obtains results exceeding 10 percentage points above the baseline. As for the approaches based on context information provided in the form of dialog act labels, it is interesting to notice that the influence pattern revealed on the Switchboard corpus and the Tilburg DialogBank dialogs is also present on the LEGO corpus, with information from the first preceding segment leading to a large increase in accuracy and information from the following leading to smaller and smaller increments. However, in this case, the improvement provided by information extracted from the first previous segment is not as high as for the n-gram-based approaches. This shows that the n-grams from the fixed system segments are able to provide more fine-grained information than the simple dialog act label. However, this can be explained by the nature of the corpus itself, as the kind of information or instruction provided by the system highly limits the possible user dialog acts. Finally, it is important to notice that, once again, information from other dimensions significantly improved accuracy by 2.21 percentage points when using a single preceding segment, but only an insignificant 0.27 percentage points on the top results.

\begin{table}[htbp]
    \caption{Accuracy (\%) results obtained on the user segments of the LEGO corpus using context information extracted from the \em n \em preceding segments in different forms. The first two rows refer to context information provided in the form of n-grams while the remaining two refer to context information provided in the form of dialog act classifications relative to the \em Task \em dimension only or all the dimensions.}
    \label{tab:legouserresults}
    \begin{tabular}{l c c c c c c}
        \hline
        & \multicolumn{6}{c}{\# Previous Segments} \tabularnewline
        & 0 & 1 & 2 & 3 & 4 & 5 \tabularnewline
        \hline
        Untagged N-Grams            & 77.25 & \textbf{88.67} & 84.53 & 79.58 & 69.90 & 73.37 \tabularnewline
        Index-Tagged N-Grams        & 77.25 & \textbf{88.87} & 88.69 & 87.91 & 87.69 & 87.12 \tabularnewline
        Manual Annotations (Task)   & 77.25 & 85.99 & 87.81 & 88.10 & \textbf{88.48} & 88.43 \tabularnewline
        Manual Annotations (All)    & 77.25 & 88.20 & 88.65 & \textbf{88.75} & 88.67 & 88.73 \tabularnewline
    \end{tabular}
\end{table}

Overall, the experiments on data annotated according to the ISO 24617-2 standard led to results and patterns similar to the ones obtained on the Switchboard corpus. This is important, since it means that our conclusions are not specific to one dialog act annotation tag set. Furthermore, except for some aspects, the conclusions are also corpora-independent. The main differences occurred on experiments on the LEGO corpus, on which there were almost no performance differences between approaches in terms of maximum accuracy. However, this was explained by the nature of the system segments that were used to provide context information. In terms of multilinguality, not many conclusions could be drawn, since the results on the Dutch dialogs obtained from the Tilburg DialogBank led to irregular patterns, probably due to the reduced amount of data. However, the importance of context information and some of the patterns were still noticeable.

%
%

%
%

\section{Discussion}
\label{sec:discussion}

In this document, we presented an analysis of the influence of context information on dialog act recognition. The analysis was performed on the widely explored Switchboard corpus, as well as on data annotated according to the recent ISO 24617-2 standard. While the first was chosen for its large amount of data and for the sake of comparison with previous research in the area, the latter was chosen in an attempt to contribute for the standardization of experiments in the area. In this sense, in addition to data obtained from the Tilburg DialogBank, we also used the LEGO corpus by mapping the original annotations of the corpus into the communicative functions of the standard. 

Context information was obtained from one up to five preceding segments and provided in three ways. The first, untagged n-grams, that is, using n-grams from previous segments in an indistinguishable way from the ones of the current segment, was generally detrimental. The only exception was on the LEGO corpus, where untagged n-grams from the first preceding segment were able to improve the baseline accuracy. However, this was due to the rigid nature of the system segments in the corpus and the approach was still outperformed by the others on the same dataset. The second approach to provide context information was in the form of index-tagged n-grams, that is, n-grams tagged with the distance between the segment they were extracted from and the current segment. In this case, accuracy highly improved using a single previous segment. However, beyond that, there were no visible improvements. Finally, information provided in the form of dialog act classifications was able to gradually improve accuracy and revealed similar influence patterns on every corpus. In this sense, the influence of preceding segments seemed to decrease exponentially with the distance. Furthermore, it is important to notice that the same patterns were verified even when using automatic annotations instead of the manual annotations of the gold standard. Also, in the case of data annotated according to the ISO 24617-2 standard, including information from dimensions other than \em Task \em led to slight accuracy improvements, up to a maximum of 2 percentage points. However, in general, these improvements were not significant.

In terms of the language independence of the conclusions, it is difficult to make any particular assessment, since we were only able to obtain a reduced amount of non-English data and, thus, the obtained results were irregular. However, the importance of context information is still highly noticeable and some of the influence patterns are still observable.

Overall, our experiments proved that context information extracted from preceding segments is able to improve classification performance on the dialog act recognition task, independently of corpora characteristics, tag sets, and language. However, that information should be provided in a  manner distinguishable from information extracted from the current segment, that is, the features representing context information should be distinct from the ones representing the current segment. Otherwise, it may have a negative effect. This distinction can be made either by tagging the features with an index relative to the segment they were extracted from, or by using different kinds of features for context and current segment information. The first preceding segment is the most informative, typically leading to the best results when using n-grams and the largest performance improvement when using dialog act classifications. In this sense, it is not recommended to use information from additional preceding segments when using n-grams, as the outcomes varied among the different corpora. On the other hand, the approach based on dialog act classifications benefits from information from additional segments, until the results start to stabilize, around the third preceding segment. Finally, in terms of overall performance, the approach based on dialog act labels typically achieved the best results, even when using automatic annotations.

Finally, in addition to the conclusions about the importance and influence of context information, it is important to notice that our experiments on the Switchboard corpus also led to results that advanced the state-of-the-art on the dialog act recognition task on that corpus, using every variant of the tag set. Furthermore, the results obtained on data annotated according to the ISO 24617-2 standard define a baseline for future work and contribute to the uniformization of experiments in the area.

%
%

%
%

\section{Future Work}
\label{sec:future}

In our experiments, we only considered textual features. However, the studies by \namecite{Wright1998}, \namecite{Stolcke2000}, \namecite{Dielmann2007}, and \namecite{Sridhar2009} show that audio features are also able to provide important information for the dialog act classification task and are not influenced by \ac{ASR} errors. Thus, it is our intention to perform further experiments, exploring the ability of acoustic-prosodic features to provide context information for dialog act recognition.

Considering \ac{ASR}, it would be interesting to analyze how \ac{WER} influences the performance of context features. Since we have manual transcriptions of the Switchboard corpus, this can be done by generating automatic transcriptions of the same dialogs and observing the differences in performance.

Furthermore, concerning the ISO 24617-2 standard, it would be interesting to perform experiments to identify communicative functions on the other dimensions and assess whether the preceding segments are able to provide important information for those dimensions as well. 

Finally, it is important to obtain more annotated data in non-English languages, so that more extensive studies concerning the language-independence of our conclusions can be performed.  

%
%

\starttwocolumn

%
%

\begin{acknowledgments}
This work was supported by national funds through \ac{FCT} with reference UID/CEC/50021/2013, by Universidade de Lisboa, and by EU-IST FP7 project SpeDial under contract number 611396.
\end{acknowledgments}

%
%

\bibliographystyle{template/fullname}
\bibliography{references}

\end{document}